# The Control Layer Behind Rule-Following Agents Trading Across Market Conditions

Trade-level evidence from four AI agents trading with live market data over six months and what to check before agents trade in the markets for you

Saad Naja[1], Nasser Odtallah[1], Jake Van-Praag[1] and Sarah May[1]

[1]Hygentra AI

# Abstract

Trading discipline often fails people. Our study tested whether four AI trading agents followed their strategies and risk rules. Offering traders the evidence and an agent evaluation toolkit to separate luck from following the rules, before giving agents meaningful autonomy or capital.

Our study analysed 2,627 closed positions from 12 February to 27 August 2026 across the same asset universe and execution conditions. Most positions were simulated against live prices with a 0.09% round-trip fee; 18% involved real Hyperliquid orders. Each position declared a strategy, stop and target, with a limit of five open positions per agent.

The fleet finished net positive after fees. Rule adherence was high across all four agents, including the unprofitable one. Profitability tracked how often agents won relative to their break-even win rates. Institutional-style controls ran by default and contained losses through a 29% market drawdown and recovery, with documented exceptions.

The discipline held for all four agents. Agents kept 97.5% of gross loss within declared stops, and typical losses closed at declared levels. Declared risk remained stable after losses and broadly stable across calm and volatile months. Twenty-three of 24 runs respected the position limit; logs identify the exception's size and duration. The same live data produced different results across assigned strategies.

The trade-log tests can be rerun from the published trade snapshot; regime tests also require the published Bitcoin prices and regime labels. Section 3.5 specifies the stop-breach tolerance. These trade-level results are not account-level returns or proof of a lasting edge. Our study provides an evaluation toolkit to help traders distinguish rule-following from luck before committing capital. For anyone considering allocating capital to agents to trade, discipline should be the entry test, not the only test.

# 1. Introduction

## 1.1 Why discipline is the most important test for AI agents

AI systems can monitor markets continuously and execute trades in milliseconds. AI systems don't change in response to emotion, fatigue, or personal exposure unlike humans. In crypto markets, where trading never stops, that speed is part of their appeal and why it's the first proving ground. Speed only tells part of the story. There's a second premise, made less often and tested less often. AI trading agents can also be superior rule followers compared to their human counterparts, precisely because controls can be programmed.

Most traders start out with the right intentions setting out their strategy and risk framework including a position size, a stop, a maximum exposure, conditions for standing aside. Writing it down isn't the hard part. Following it in the moment is the hard part. Humans often abandon their own rules when their emotions take over. Our prior longitudinal study of human traders identified the 27-year pattern behind their failure rate, between 74–89% of retail traders still lose money, losing for the same behavioral reasons every time: overconfidence, revenge trades, fatigue, bad timing, in the same four-phase spiral no matter the market cycle, or trading education **(Section 8. Related Work)**. Discipline is the aspect of trading that people are worst at and that programmable AI can be best at.

If AI agents can automate trading strategies with discipline this premise is more consequential than just speed and market access. AI trading agents can do more than execute the trades a person prompts. That matters, because the trades people choose lose about four times out of five **(Section 8. Related Work)**.

It brings the institutional playbook within reach of retail investors. Professional trading desks use more than a market view. They add pre-trade checks, position limits, stop rules, risk reviews, and procedures for unusual conditions. Those controls don't guarantee a profit, but they make the system's behavior easier to inspect and its losses easier to contain. A retail investor may not have institutional capital or market access, but an AI trading agent can help apply some of the same habits into real, enforceable controls.

## 1.2 Why the evidence is missing

A trader trying to assess whether a given AI agent actually follows its trading strategy and rules has very little evidence to make an informed decision. The idealist would say this lack of data is expected for where a nascent breakthrough industry is in its cycle, the pragmatist would take their time to offer a considered explanation why, show their cards and push for higher transparency standards.

**Trust in LLM benchmarks is low, and mostly measures horsepower.** Trust in AI benchmarks is low because models can game standardized tests, tests fail to reflect messy real-world tasks, and evaluations often ignore hidden variables like test-time compute or prompt harnesses (PCMag). The comparative literature evaluates which model trades best, and has converged on a consistent answer: most LLM agents fail to beat buy-and-hold on equities over a multi-month horizon Chen, Y., et al. (2025). *StockBench: Can LLM Agents Trade Stocks Profitably In Real-world Markets?* arXiv:2510.02209 [cs.LG]. General intelligence does not automatically translate into effective trading capability across US stocks, A-shares or crypto Fan, T., et al. (2025). *AI-Trader: Benchmarking Autonomous Agents in Real-Time Financial Markets.* arXiv:2512.10971 [q-fin.CP]. System-level design choices influence profitability more than the choice of model Qian, L., et al. (2025) *When Agents Trade: Live Multi-Market Trading Benchmark for LLM Agents.* arXiv:2510.11695 [cs.CL]. This work indicates the model isn't the binding constraint. At the time of this report, we couldn't identify evaluations that report the brakes: pre-trade checks, live risk review, hard position limits, and whether any of them held.

**Few are questioning the exchange (house) wins mentality.** Existing AI trading features offered by trading apps and exchanges have offered limited autonomy to agents putting the trading strategy burden on the user re-routing a trade, which based on our longitudinal research is destined to lose four times out of five (74–89%) **(Section 8. Related Work)**, now faster with AI. The first wave of retail-facing tools automates an instruction the user prompts: buy this asset e.g Bitcoin between these prices, exit at that level. It means the agent isn't accountable for the strategy, only for carrying it out, and nothing in the design asks whether the human's trade was a profitable one. The novel capability being sold is execution because it drives trading volumes. Institutional desks stay solvent because their controls have teeth. Pre-trade compliance runs before every order, hard risk limits sit above every book, and the system blocks allocation or concentration breaches instead of flagging them afterwards. None of that depends on the trader's emotion. The quality of the AI system depends on the quality of the trading process encoded into it. Developers have to ground an agent's trading logic in a structured allocation process and give it rules shaped by real market experience from trading desks.

**The agent token market has been measured, not trading agent's performance.** The most systematic audit of the category to date surveyed more than 1,900 AI-tagged crypto projects and analysed eleven agent treasuries covering 925,323 token holders (Yu, J., Zhao, A., and Sui, D. (2026). *Paper Agents, Paper Gains: An Empirical Analysis of DeFi Investment Agents.* arXiv:2605.29174 [cs.AI]. Its headline numbers are severe — aggregate holder gains peaked above $2.4 billion before collapsing to a net loss of $191.7 million, with the top 1% of profitable wallets taking 81.4% of all gains but its data about a token market. Many sampled deployments lacked clear evidence of autonomous execution. Holder returns turn on entry timing, token liquidity and secondary demand for a project's coin; treasury figures turn on whatever assets a treasury happens to hold. The authors are explicit that treasury performance does not pass through to holder returns in any reliable way. Neither quantity says whether any agent traded competently.

**Lack of industry-wide agent performance.** Evidence on agent trading reaches the public from three directions, none of which yields a benchmark. Companies and projects release partial figures, shaped by a nascent product and by what they're prepared to disclose. Researchers analyse third-party data without the operational context that would make it interpretable: the mandate, the human intervention points, what the execution layer was doing. Individual hobbyist influencers publish selected results, where survivorship does the work. Each produces a narrative; none produces a figure a third party can recompute. **Section 8.** reviews this literature. It also applies to this paper, which is an agentic trading lab's paper about its own agents: which is why every figure here traces to a hashed snapshot, the roster rule is stated as executable logic, and the specifications that weaken our results are reported beside those that do not.

**Offering an agent evaluation toolkit and calling for more collective transparency** As more agents reach the market, the problem facing a prospective trader is agent selection: which of these to allocate capital to, and on what evidence. That doesn't mean placing the cognitive burden on people to build an agent from scratch. Without decades of trading grounding and robust data sets, lucky trading can pass for competence while someone else's money carries the risk. For trading AI agents to succeed, measurement must be fit-for-purpose and transparent otherwise benchmarks serve narratives not verifiable facts. Following the same mistakes that have resulted in distrust of LLMs benchmarks. By sharing a snapshot of our agent performance findings, we offer up a practical toolkit for retail investors to shortcut agent selection and test which claims can be verified rather than taken on trust. For the eager skimmers before allocating capital, head to **Section 6. A trading agent evaluation toolkit** for the tests we recommend to evaluate AI trading agents before allocating capital.

## 1.3 What this paper measures

This paper measures whether agents can be disciplined, whether that discipline is visible from the outside, and what actually separates profitable agents from unprofitable agents: using agent performance data rather than a generic model benchmark.

The evidence base is a fleet of four independent agents, each with a distinct trading strategy mandate, run under identical conditions on the same asset universe for 6.5 months. The fleet finished net positive after fees: +0.050% per trade pooled and +0.221% with agents weighted equally. Per-trade results ranged from clearly positive to slightly negative. Agent-2 averaged +0.836% and Agent-4 +0.062%, Agent-3 was close to flat at +0.028%, and Agent-1 averaged −0.043%. These figures include every recorded trade. **Section 7.2** shows how they change when the August outage trades are excluded. That dispersion is the object of study. We are not measuring how good these four agents are; we are testing what separates a profitable agent from an unprofitable agent, using a fleet where the answer is already known and the trade-level record is complete.

## 1.4 Research questions

Five questions were reported on:

1. Rule adherence: Did the agents follow the trading rules that were programmed, and can we clearly see when agents didn't?
2. Attribution drivers: rule adherence vs asset mix: Is agent performance attributable to rule adherence rather than exposure to a few favourable assets?
3. Attribution drivers: hit rate vs payoff ratio: Are agents more profitable because they earn more on winning trades or because they win more often?

4. Regime fit: Do agents perform better when they trade in the same direction as the current market trend?
5. Strategy Rotation: Do specific agents perform better during the same type of market conditions?

## 1.5 Contributions

This research paper makes six contributions:

1. It provides a fully specified trade-level account of what separated profitable from unprofitable agents across four common trading strategies and 2,627 closed paper positions, including the complete roster, autonomy boundary, mechanical inclusion rules, and pooled, equal-weighted and ex-Agent-1 results **(Sections 2.1–2.5 and 3.1–3.4)**.

2. It introduces a nine-test audit battery that a third party can reproduce from a trade log: five tests of loss containment, parameter stability, behavior after losses and position limits, followed by four tests separating agent conduct from asset selection **(Sections 4.1–4.2)**.

3. It decomposes expectancy to show that the agents' profitability ranking tracked the margin between their observed hit rate and strategy-specific break-even hit rate, rather than payoff ratio, loss size or asset mix, and tests that result through counterfactual and loss-capping analyses **(Sections 4.2–4.3)**.

4. It constructs hindsight-free regime labels from benchmark prices using an expanding window, then tests directional alignment, volatility sensitivity and weekly ranking stability across changing market conditions **(Sections 3.3 and 4.4–4.5).**

5. It reports uncertainty and robustness explicitly through day-block bootstrap intervals, day-clustered regression errors, multiple-comparison correction, alternative fleet weightings and three treatments of the August execution outage, while separating conclusions supported by simulated trade-level data from account-level or live-execution claims the dataset cannot support **(Sections 3.4 and 7).**

6. It translates the findings into implications for agent evaluation and an eight-step due-diligence toolkit organised around three criteria: whether an agent's process is visible, disciplined and tested across conditions **(Sections 5.1–5.2 and 6.1–6.8).**

# 2. Agents and their trading environment

## 2.1 Roster of Agents

Hygentra.ai is an AI trading company aiming to onboard retail investors and brokers into agentic trading via its agentic trading platform. Self-directed traders can ask an agent to read a chart, review an order, explain the setup or run a named strategy inside the limits traders set. Hygentra gives traders an AI-powered workspace designed around individual trading behavior, better execution, and delegating trading decisions to AI agents with visible controls. Making agentic trading easy to understand, easier to supervise, and easier to trust. Hygentra is for active traders who want help making faster, clearer trading decisions without giving up control. It is useful if you want better chart context, cleaner trade review, agent-led execution, or a way to understand your own trading patterns.

Any broker can license Hygentra's intelligence layer to offer hedge-fund-grade AI to their clients, giving brokers and exchanges agentic infrastructure they would otherwise spend years building in-house. Hygentra was founded by Saad Naja with a team of former investment-bank executives, quants and day traders. Naja brings extensive experience from building trading products across trade, institutional and retail at Goldman Sachs, Deutsche Bank and Kraken. For the last 6.5 months, Hygentra has been setting rules and trading strategies for four agents across changing market conditions. AI Agents are named by trading strategies with a specific job.

Hygentra's four agents reviewed had different strategic mandates:

1. Agent-1, a high-tempo momentum trader that takes small, short-held positions
2. Agent-2, an interday agent with a reversal mandate, multi-day holds and the widest stops in the fleet
3. Agent-3, a selective breakout trader that sits out the setups that do not qualify and runs low leverage; and
4. Agent-4, a trend follower.

For the purpose of introducing the agents and their trading history, the next two tables provide a summary snapshot of performance by agent. **Section 4. Findings** provide detailed research findings on what factors drove agent profitability, and **Section 5. Discussion** provides inferences and interpretations of discipline hypotheses. **Section 6. A trading agent evaluation toolkit** offering interested parties tests.

**Table 2.1a** shows an agent named Agent-1 dominates the sample: Agent-1 accounts for 1,507 of 2,627 trades (57%). Any fleet-aggregate figure is substantially a statement about Agent-1. We report equal-weighted and ex-Agent-1 aggregates alongside in **Table 2.1b.** On survivorship: all four were active at the cut date, no agents were retired during the window

**Table 2.1a — Agent performance snapshot, 12 February to 27 August 2026**

| Agent | Strategy | Class | Closed trades | First trade | Last trade | Active days | Median hold | Median declared stop | Median target/stop | Short share |
|---|---|---|---|---|---|---|---|---|---|---|
| **Agent-3** | Breakout | Day | 349 | 12 Feb | 27 Aug | 72 | 4.3 h | 1.0% | 2.5x | 48.0% |
| **Agent-2** | Reversal | Interday | 178 | 12 Feb | 26 Aug | 82 | 57.1 h | 3.9% | 2.0x | 76.0% |
| **Agent-4** | Trend-following | Day | 593 | 13 Feb | 27 Aug | 109 | 5.0 h | 1.0% | 2.0x | 58.0% |
| **Agent-1** | Momentum | Day | 1,507 | 13 Feb | 27 Aug | 163 | 2.9 h | 0.8% | 2.5x | 58.0% |

*Note:* All definitions can be found in one place within **Section 3. Methodology and Terminology**

**Table 2.1b** shows the trade-level series. Agent-1 accounts for 1,507 of 2,627 positions, or approximately 57% of all trades. Any fleet aggregate is therefore substantially a statement about Agent-1's activity. The agent-level table is more informative than a single fleet total.

**Table 2.1b — Agent trade-level results**

| Agent | Closed trades | Hit rate | Avg win | Avg loss | Median win | Median loss | Payoff | Expectancy | Profit factor | Sum of returns (pp) |
|---|---|---|---|---|---|---|---|---|---|---|
| Agent-2 | 178 | 55.1% | 5.020% | 4.300% | 5.760% | 4.010% | 1.170 | +0.836% | 1.430 | +148.8 |
| Agent-4 | 593 | 50.8% | 1.200% | 1.110% | 1.480% | 1.090% | 1.080 | +0.062% | 1.110 | +37.0 |
| Agent-3 | 349 | 42.7% | 1.670% | 1.200% | 1.350% | 1.090% | 1.400 | +0.028% | 1.040 | +9.9 |
| Agent-1 | 1,507 | 42.5% | 1.140% | 0.920% | 1.310% | 0.890% | 1.240 | −0.043% | 0.920 | −64.9 |
| **Fleet, pooled** | **2,627** | **45.3%** | **1.540%** | **1.182%** | **1.462%** | **0.924%** | **1.303** | **+0.050%** | **1.077** | **+130.9** |
| **Equal-weighted agents** | **2,627 †** | **47.8%** | **2.258%** | **1.879%** | **2.475%** | **1.772%** | **1.222** | **+0.221%** | **1.127** | **+32.7 †** |
| **Ex-Agent-1, pooled** | **1,120** | **48.9%** | **2.010%** | **1.584%** | **1.610%** | **1.090%** | **1.269** | **+0.175%** | **1.216** | **+195.7** |

*Note:* Fleet and ex-Agent-1 figures pool the underlying trades. Equal-weighted figures are the simple mean of the four agent-level statistics, giving every agent the same weight regardless of trade count. †All 2,627 trades are represented, but the agents are weighted equally. The +32.7 percentage-point figure is the mean of the four agents' summed trade returns. None of the summed-return figures represents a compounded account return.

Agent-2's sum of returns "+148.8 pp" means that the 178 trade unweighted net-return percentages sum to 1.48798812 in decimal form; it does not mean that a funded account grew by 148.8%.

Expectancy is mean net return per trade. "Sum of returns" is the sum of per-trade percentage returns in percentage points; it is not an account return.

The fleet is positive after fees under every weighting. Two agents have positive expectancy, one is close to flat and one is slightly negative. The fleet-aggregate expectancy is +0.050% per trade, the equal-weighted mean of agent expectancies is +0.221%, and the ex-Agent-1 aggregate is +0.175%.

## 2.2 Level of autonomy

Each agent uses a hosted model driven by a database prompt identified by `[strategy_name]`. The prompt configures its strategy and risk rules: with an opening stop, profit target, partial-exit ladder, and trailing stop on nearly every position, plus a maximum of five concurrent positions and a 'normal' risk-appetite setting. The model outputs the trade parameters within that framework, and the engine executes entries and exits. During a request, a tool can return past trade instructions with their outcomes and P&L, and the prompt tells the model to use that record. This is retrieval within the request. The model outputs the symbol, direction, stop, and target within the configured framework, while the engine executes entries and exits. Hygentra doesn't train, fine-tune, or update model weights as these agents trade.

Humans design the frameworks. The AI systems execute within them. Nothing in this paper should be read as a claim of unsupervised autonomy.

Human intervention points, as visible in the data:

- **Configuration.** Between 13 and 25 August, engine-side stop-loss and take-profit execution was disabled by a configuration flag during an infrastructure migration. This was a human/infrastructure event, not an agent decision.
- **Reconciliation.** 323 of 2,627 exits (12.3%) were closed by reconciliation jobs, external venue events or run termination rather than by the agent's own stop, target or trailing stop. Two further exits are labelled as manual ledger repairs.
- **Trading runs.** From 9 May, each agent was instantiated in a second concurrent trading run alongside the original. 31% of second-run positions were opened while the same agent held an open position in the same symbol and direction in the first run. The runs aren't mirrors. Most model outputs differ, but overlapping positions reduce the effective number of independent trade instructions below the trade count.

## 2.3 Assets agents traded

The declared universe contains twelve instruments: eight crypto perpetuals (BTC, ETH, SOL, BNB, LINK, DOGE, ASTER and HYPE) and four blockchain-traded commodity perpetuals (gold, silver, WTI and Brent). The trade log additionally contains 165 positions in XRP and LTC, which are not in the declared universe. We keep them in the primary specification and remove them as a robustness check; no result changes.

Concentration is high: BTC, ETH, SOL, LINK, DOGE and BNB account for 89% of trades. Commodities account for 59 trades (2.2%).

## 2.4 Market conditions over the window

The window is short but not quiet, and reading the results without the market backdrop will mislead.

**Table 2.4 — Market conditions**

| Month | BTC month-end | Return | Dominant regime days |
|---|---|---|---|
| Feb 2026 | 66,973 | — | DOWN_HIGH (20) |
| Mar | 68,284 | +2.000% | DOWN_MID (13), UP_MID (10) |
| Apr | 76,347 | +11.800% | UP_LOW (26) |
| May | 73,674 | -3.500% | DOWN_LOW (19), UP_LOW (12) |
| Jun | 58,625 | -20.400% | DOWN_MID (20) |
| Jul | 62,888 | +7.300% | UP_LOW (19) |
| Aug | 80,250 | +27.600% | DOWN_LOW (10), UP_LOW (8), UP_MID (6) |

BTC closed the window 4.3% above its 1 February close — a number that conceals almost everything that happened in between. Peak to trough was −28.7%, from 82,210 on 10 May to 58,625 on 30 June. Realised volatility annualised at 47.5%. The largest single-day move was −14.0% (5 February); the second and third largest were both upward and both in the final fortnight (+7.1% on 19 August, +7.3% on 21 August).

Two features of this backdrop matter for interpretation. First, the fleet's most common market state was the low-volatility uptrend (703 trades, 27%) — the state in which every agent lost money. Second, a full 6.5-month window still contains only one cycle: a drawdown followed by a recovery. An agent with a short bias will look skilled in the first half and unskilled in the second, and we cannot separate the two explanations at this sample size, see **Section 7.6** for sample generalisation limits and **Section 4.4 Table 4.4b** for regime results.

For retail investors. The 2,627 trade count measures how often an agent acted, not how many different market conditions it survived — an agent can log thousands of trades inside a single market regime. These agents met high-volatility conditions on only 34 of the window's 208 days, and the record as a whole contains one drawdown followed by one recovery. So ask an agentic trading lab for the calendar rather than the trade count: how many days of each market condition the track record actually covers. Where a condition is thin or absent, the agent is untested in it, which is not the same as having passed.

Activity wasn't continuous. Agent-3 closed 9 positions in the eleven weeks from 9 March to 24 May. Agent-4 closed 14 in the eight weeks from 2 March to 26 April. Agent-2 closed none in April. Agent-2 opened its last new position on 11 August and closed its last on 26 August. It wasn't trading new positions during the 19–21 August market move. Agent-1 traded throughout. Cross-agent comparisons therefore compare agents that were exposed to somewhat different sub-windows, and we test this in **Section 6.**

## 2.5 Simulated and live trades

From 12 February to 20 July 2026, the agent positions are simulated paper positions mirrored against live venue prices, not a live order book. The supplied net-return field applies a 0.09% round-trip taker fee assumption, equivalent to 0.045% per side. Funding costs are not included because they are not reliably recorded per trade.

The trade unit is one discrete position from open to close on one agent and one symbol. It isn't a FIFO round-trip matching. Partial exits are represented as a reduced close on the same row rather than a separate round trip. These choices make the trade log usable for behavioral decomposition but do not turn it into an account-level performance series.

From 21 July 2026 to 27 August 2026, 483 of the 2,627 trades (18.4%) placed were real orders on Hyperliquid. Across the full period, the dataset combines 2,144 simulated paper positions with 483 real Hyperliquid orders, equal to 18.4% of all trades.

# 3. Methodology and Terminology

## 3.1 Trade unit and return

| | |
|---|---|
| A trade | is one position from open to close on a single agent and symbol, as recorded by the execution engine. Rows are included only where \`is_open = false\`, \`close_date IS NOT NULL\`, \`open_rate > 0\` and \`close_rate > 0\`. |
| Gross return | = ((close_rate − open_rate) / open_rate) × (−1 if short else +1) |
| Net return | = gross_return − 0.0009 |
| Return metrics | All returns are percentage returns on entry price. We re-derived `net_return` from the price columns for every row; the maximum discrepancy is $5 \times 10^{-9}$. |

## 3.2 Per-agent metrics

| | |
|---|---|
| Hit rate | = winning trades ÷ all trades, on `net_return > 0` |
| Avg win | = mean of positive `net_return`; `avg_loss` = mean of \|negative `net_return` |
| Medians wins | are reported alongside because the win distribution is right-skewed. |
| Payoff ratio | = `avg_win ÷ avg_loss` |
| Profit factor | = gross wins ÷ gross losses |
| Total return | = sum of `net_return` across all trades. A sum of per-trade percentage returns, not a compounded portfolio return. |
| Expectancy decomposition | `cf_own_payoff_median_hit` substitutes the fleet-median hit rate into the agent's own win/loss sizes. `cf_own_hit_median_payoff` substitutes the fleet-median payoff ratio while keeping the agent's own hit rate and loss size. Whichever counterfactual better reproduces the observed profitability ordering identifies the dominant driver. We add a loss-normalised version (expectancy in units of the agent's own average loss, "R"), because Agent-2's per-trade returns are roughly four times larger than the day-traders' and the raw counterfactuals mix scale with structure. |
| Discipline | Discipline is examined through holding-time consistency, exit composition, and stop behavior. Loss-containment metrics compare gross loss with the separately stored opening stop. A later stop value can differ because the trailing stop moved during the trade. That movement records the trail working as configured. `stop_breach_gross` is the share of all losing trades whose gross loss exceeded the declared stop, compared gross because the declared stop is a gross price level. It measures outcome discipline including exits that never touched the stop. Stop execution quality is measured separately on `exit_reason = 'SL'`. `declared_max_hold` does not exist as a field, so the brief's |
| | mandate-breach rate is not computable; hold-time dispersion (`hold_median_h`, `hold_iqr_h`, `hold_cv`) is reported instead. |

## 3.3 Regime labels

| | |
|---|---|
| Regime | Labels are built from BTC daily closes only. No agent data was joined at any point during construction, and the BTC source file and frozen regime-label file are published with the report. |
| Direction | = sign of the trailing 7-day BTC log return |
| Realized_vol_30d_annualised | = standard deviation of the trailing 30 daily log returns × √365 |
| Vol tercile | = LOW / MID / HIGH against an expanding distribution of all prior observations — never the full-sample distribution |
| The expanding window | is what makes the label hindsight-free: every label uses only data available on its own date. The first day was `WARMUP` and excluded. Over 208 days: UP_LOW 65, DOWN_LOW 47, DOWN_MID 36, UP_MID 25, DOWN_HIGH 21, UP_HIGH 13. |
| Join rule | A day's label is computed from that day's close. To avoid using a close that had not yet printed when a position was opened, each trade is joined to the label of the previous calendar day. All 2,627 trades match. |
| Alignment | A trade is aligned if long in an UP state or short in a DOWN state. |
| Active days | Active are distinct calendar days on which the agent opened at least one position. The opening stop is stored separately and is present on all 2,627 rows. Median declared stop and target/stop ratio use that opening value across all included trades |

## 3.4 Inference

Trades within a day are not independent. Agents hold up to five concurrent positions, and from May two runs traded in parallel. We report trade-level tests (as most agent-performance claims do) and, alongside, day-block bootstrap confidence intervals and day-clustered regression standard errors. Where they disagree, the day-clustered result is the one to believe. The nine hypothesis tests reported in **Section 4** are corrected for multiple comparisons with the Holm procedure **(Table 3.5c).**

Specifications, in full: per Wilson (1927), *JASA* 22(158), 209–212 (source) the Wilson intervals for hit rates; $\chi^2$ for hit-rate differences across agents; one-sample t and Wilcoxon for expectancy; Mann–Whitney

U for two-group return comparisons; Kruskal–Wallis for multi-group. The day-block bootstrap resamples calendar days with replacement (4,000 draws) and recomputes trade-weighted expectancy. Regressions are OLS of `net_return` on agent, symbol, regime and side indicators, with standard errors clustered by open date. Rank-rotation periods are calendar weeks (Mon–Sun) by close date; an agent is ranked if it closed at least 3 trades and a week is ranked if at least 3 agents qualify, with the week labelled by its modal prior-day regime. The loss-capping counterfactual is `max(gross_return, −|stop|)` on losing trades.

The analysis code, frozen trade snapshot, BTC source file, and regime-label file are published with the report. All figures were regenerated from the raw trade rows and reconciled to tables exactly.

## 3.5 Discipline and asset-attribution metrics

Unless stated otherwise, all metrics are calculated on the included closed trades defined in **Section 3.1.** Returns are calculated as decimals and displayed as percentages. "pp" means percentage points.

**Table 3.5a — Descriptive, performance and aggregation metrics**

| Metric | Definition |
| --- | --- |
| Closed trades | Number of rows meeting the inclusion rules in **Section 3.1.** |
| First and last trade | Earliest and latest `open_date` among the included trades for that agent. |
| Holding time | `close_date - open_date`, expressed in hours. |
| Hold median | Median holding time across the relevant trades. |
| Active days | Number of distinct calendar dates on which the agent opened at least one included position. |
| Short share | Number of short positions divided by all included positions. |
| Median declared stop | Median absolute opening stop across all included trades. The opening stop is stored separately and present on every row. |
| Median target/stop | Median of `aimed_take_profit_pct` divided by the separately stored opening stop across all included trades. |
| Median win and median loss | Median positive net return and median absolute negative net return, respectively. |
| Expectancy | Mean net return per trade: `hit rate × average win - (1 - hit rate) × average loss`. Average loss is expressed as a positive magnitude. |
| Break-even hit rate | `average loss ÷ (average win + average loss)`, equivalently `1 ÷ (1 + payoff ratio)`. |
| Margin over break-even | Observed hit rate minus break-even hit rate, reported in percentage points. |
| Sum of returns | Sum of the net returns on the included trades, reported in percentage points. It is not a compounded or account-level return. |
| Fleet, pooled | Recalculate the metric after pooling all included trades across the four agents. Each trade receives equal weight. |
| Equal-weighted agents | Arithmetic mean of the four agent-level metrics, so every agent receives equal weight regardless of trade count. For summed returns, this is the mean of the four agent sums. |
| Ex-Agent-1, pooled | Recalculate the pooled metric after removing every Agent-1 trade. |
| BTC month-end | Last available BTC daily close in the calendar month. |
| Monthly BTC return | Current month-end BTC close divided by the previous month-end close, minus one. February is left blank because the study begins during that month. |
| Dominant regime days | Regime labels with the largest calendar-day counts within the month. The number in parentheses is the number of days assigned that label. |
| Aligned-minus-misaligned difference | Aligned-trade expectancy minus misaligned-trade expectancy, reported in percentage points. |
| By-volatility result | All UP and DOWN regime cells within the same LOW, MID or HIGH volatility tercile pooled before expectancy is recalculated. |

**Table 3.5b — Discipline and asset-attribution metrics**

| Metric | Definition |
|---|---|
| Declared-risk multiple | For each losing trade, using the separately stored opening stop, `R-stop = gross-loss magnitude ÷ absolute declared stop`. A value of 1.00 means the loss equals the position's declared stop. |
| Loss contained inside the declared budget | `1 - [sum of loss beyond each stop ÷ total gross loss]`, where loss beyond the stop is `max(gross-loss magnitude - absolute stop, 0)`. |
| 95th-percentile loss in declared risk | Ninety-fifth percentile of `R-stop` across all losing trades, using the separately stored opening stop. Quantiles use linear interpolation between order statistics (the NumPy default). |
| Parameter stability | Median absolute declared stop within each prior-day volatility tercile. **Table 4.1a** compares HIGH with LOW. |
| Risk after wins and losses | Median absolute stop on trades opened after a win, after one loss and after two consecutive losses within the same agent-run. Only outcomes known when the next trade opened should determine the sequence. |
| Post-loss trading frequency | Share of eligible trades opened after the agent's most recently completed trade was a loss. |
| Expected post-loss frequency | Agent's unconditional net-loss rate. This is the expected post-loss share if the previous outcome does not affect trading frequency. |
| Agent-run | One agent within one `trading_run_id`. Runs are assessed separately before results are counted across the agent. |
| Concurrent positions | Number of positions open within the same agent-run at a given timestamp. A position counts from its open timestamp until, but not including, its close timestamp. |
| Position-limit result | Number of agent-runs in which maximum concurrency never exceeded five, divided by all observed agent-runs. |
| Hold IQR | Seventy-fifth percentile holding time minus twenty-fifth percentile holding time. State the quantile interpolation method used. |
| Hold CV | Population standard deviation of holding time divided by mean holding time. |
| Stop exits | Trades with `exit_reason = SL`, divided by all included trades for the agent. |
| Target exits | Trades with `exit_reason = TP`, divided by all included trades for the agent. |
| Breakeven-trail exits | Trades with `exit_reason = TSL_HIT_AT_BREAKEVEN`, divided by all included trades for the agent. |
| External or reconciled exits | Trades labelled `EXTERNAL_HL_CLOSE`, `RECONCILED_HL_POSITION_GONE`, `RECONCILED_VENUE_TRUTH_BACKFILL` or `TRADING_RUN_CLOSED`, divided by all included trades. |
| Other exits | Trades labelled `PARTIAL_TP` or as a manual ledger repair, divided by all included trades. These rows should either be shown as a separate **Table 4.2c** column or explicitly identified as excluded from the four named exit categories. |
| Target share of stop and target exits | `TP exits ÷ (TP exits + SL exits)`. Other exit types are excluded from the denominator. |
| `stop_breach_gross` | Share of gross-losing trades whose gross loss exceeds the absolute opening stop by more than 0.05 percentage points. The tolerance is $\varepsilon = 0.0005$ when returns use decimal units. This tolerance sits just above the 90th percentile of normal fill variance. |
| Positive-contribution concentration | Largest positive symbol-level sum of net returns divided by the sum of all positive symbol-level contributions for that agent. |
| Common-mix reweighting | For each agent, calculate mean net return by symbol and apply one common set of fleet symbol weights. Where an agent never traded a symbol, that cell is excluded and the remaining fleet weights are renormalised to sum to one |
| Leave-one-asset-out result | Recalculate each agent's summed net return after removing every trade in one symbol, repeating the calculation once for each symbol. |
| Statistical asset controls | OLS coefficient for each agent after controlling for symbols. The expanded specification additionally controls for prior-day regime and trade direction, with standard errors clustered by open date. |
| Adherence verdict | Verdict from comparing each adherence metric with a stated threshold. Pass: (1) at least 95% of gross loss inside the declared stop and a 95th-percentile loss of 1.25 R or less; (2) median declared stop in high volatility within 10% of low volatility; (3) median stop after one and two losses no wider than after a win; (4) post-loss trade share within 2 percentage points of the loss rate; (5) no agent-run above five concurrent positions. Partial: one agent or one run misses the threshold. Fail: two or |
| | more miss it. Readers can apply stricter thresholds from the published snapshot. |

**Table 3.5c — Ranking and inference metrics**

| Metric | Definition |
|---|---|
| Rank | Agents ordered from highest to lowest expectancy. Ties receive average ranks. |
| Spearman ρ versus observed | Spearman rank correlation between the four agents' observed expectancy ranking and the ranking produced by the relevant counterfactual. |
| Loss-normalised expectancy | Expectancy divided by the agent's average loss magnitude. Report this as `expectancy/average loss`, not "R", to distinguish it from the declared-stop risk multiple used in **Table 4.1a.** |
| Hit-rate interval | Two-sided 95% Wilson confidence interval for the proportion of trades with net return greater than zero. |
| Regime-cell expectancy | Mean net return among trades assigned to the same prior-day direction and volatility label. |
| Weekly agent rank | Agents ranked by mean net return in an eligible calendar week. |
| Eligible agent-week | An agent-week containing at least three closed trades. |
| Rankable week | Calendar week in which at least three agents are eligible. The week receives the modal prior-day regime label among its trades. |
| Consecutive-week pair | Two adjacent rankable calendar weeks. "Same regime", "different regime" and "same volatility tercile" are assigned from their weekly labels. |
| Top agent changed | Indicator equal to one when the highest-ranked agent differs between the two weeks in a pair. |
| Mean Spearman ρ between weeks | Mean rank correlation across the qualifying consecutive-week pairs, calculated among agents ranked in both weeks. |
| Random-ranking benchmark | Expected probability that the top agent changes under independent random rankings: `1 - 1/k`, where `k` is the number of agents compared in that pair, averaged across the observed pairs. |
| Reported p-value | Two-sided p-value unless a directional alternative was specified before analysis. The significance threshold is 0.05. |
| Holm correction | Per Holm (1979), *Scandinavian Journal of Statistics* 6(2), 65–70 (source), Holm adjustment across the nine hypothesis tests reported in **Section 4:** hit rate across agents ($\chi^2$, p = 0.0002), symbol effects (p = 0.0004), returns across regime cells (Kruskal–Wallis, p = 0.002), Agent-4 by volatility tercile (p = 0.003), Agent-1 by direction (p = 0.009), Agent-2 versus Agent-1 with symbol controls (p = 0.08) and with regime and direction controls (p = 0.10), aligned versus misaligned trades (p = 0.836), and weekly leadership in same-regime pairs (binomial p = 0.55). The first five survive correction. |

# 4. Findings

## 4.1 Rule adherence

***Did the agents follow the trading rules that were programmed, and can we clearly see when agents didn't?***

Key finding

Adherence to trading rules was high across the fleet and measurable: 97.5% of loss inside declared budgets, parameters that did not drift or widen under pressure, no escalation after losses, and a single dated position-limit breach. Five separate tests of whether an agent followed its own declared parameters returned three clean passes and two partial passes. Both exceptions are dated and bounded. All four agents followed their observable execution discipline around stop placement, rule-based exits, and winner-versus-loser holding times. Most importantly this dataset makes the exceptions identifiable and auditable.

The takeaway

A credible and transparent agent evaluation should publish the full stop-exit distribution, separate rule-based exits from external or reconciliation closes, show winner-versus-loser holding times, and disclose any period when normal controls were disabled. Retail investors considering allocating capital to trading agents should ask how the system handles losing trades. In this analysis, declared stops generally filled at the intended level and large execution overshoots were uncommon.

An agent that passes our five adherence tests has demonstrated that its trading rules system works, which is where trading agent evaluation should start rather than where it should finish. Adherence to trading rules doesn't necessarily guarantee agent profitability. The next finding, **Section 4.2** Attribution drivers: rule adherence vs asset mix begins, to unpack agent performance attribution further.

Analysis: Five adherence tests

In five separate tests of whether an agent followed its own declared parameters returned three clean passes and two partial passes. Both exceptions are dated and bounded.

Each agent declares a strategy class, an asset universe, an opening stop, a profit target, a partial-exit ladder, and a trailing stop on nearly every position, plus a maximum of five concurrent positions. Those declarations can be checked against the trade log without knowing anything about the model. We ran five tests.

**Table 4.1a — Adherence tests against each agent's own declared parameters**

| # | Test | Measure | Agent-2 | Agent-4 | Agent-3 | Agent-1 | Verdict |
|---|---|---|---|---|---|---|---|
| 1a | Loss containment | Share of total gross loss falling inside the declared stop | 99.6% | 99.1% | 98.1% | 95.7% | Pass |
| 1b | Loss containment | 95th-percentile loss, in units of the position's own declared risk (R) | 1.03 R | 1.02 R | 1.08 R | 1.14 R | Pass |
| 2 | Parameter stability | Median declared stop, high-volatility state vs low | 3.9% / 5.0% | 1.0% / 1.0% | 1.0% / 1.0% | 0.8% / 0.8% | Partial |
| 3 | No risk escalation after losses | Median declared stop after a win / a loss / two losses | 3.9 / 3.9 / 3.9 | 1.0 / 1.0 / 1.0 | 1.25 / 1.0 / 1.0 | 0.8 / 0.8 / 0.8 | Pass |
| 4 | No frequency escalation after losses | Share of trades following a loss, observed vs expected under independence | .449 / .449 | .492 / .492 | .570 / .573 | .574 / .575 | Pass |
| 5 | Position limit (max 5) | Agent-runs in which concurrent positions never exceeded five | 6 of 7 | 7 of 7 | 3 of 3 | 7 of 7 | Partial. |

Test 1 — Loss containment. *Does a loss stay within the risk the agent declared for that position?*

Yes for 97.5% of all loss, and the typical loss lands exactly on the declared stop.

Expressing every loss in units of its own declared risk — one R being a loss exactly the size of that position's stop — the median losing trade across the fleet closes at 1.00 R. Only 1.5% of losses exceed 1.5 R and 0.8% exceed 2 R. In aggregate, 97.5% of all gross loss fell inside the declared budget. The 2.5% that escaped is 38.7 percentage points out of 1,570.5, and Agent-1 accounts for 30.5 of those 38.7 — it is the loosest of the four on this test, and the only one where more than 2% of losses ran past 1.5 R.

Test 2 — Parameter stability. *Does declared risk drift as conditions change?*

Partial, no drift for three of the four agents, and two dated exceptions that both run the safe direction.

Each agent's declared stop is close to a constant: the within-agent coefficient of variation runs from 0.12 (Agent-2) to 0.23 (Agent-3), and the median is identical across all three volatility states for Agent-4, Agent-3 and Agent-1, and across six of the seven months for all four. Two exceptions, both stated: Agent-2's median stop is 3.9% in high volatility against 5.0% in low — tighter when the market is moving, which is the opposite of the drift that would matter — and in August Agent-2 moved to 5.0% and Agent-4 to 1.5%. Neither exception is an agent loosening its risk while positions moved against it, which is the failure this test is looking for, but both are changes and we count the test as partial rather than clean.

Test 3 — No risk escalation after losses. *Does the agent take more risk to win it back?*

No agents widened its risk after losing, and one tightened.

The median declared stop after a loss, and after two consecutive losses, is identical to the stop after a win for every agent. Agent-3 in fact tightens, from 1.25% after a win to 1.0% after a loss.

Test 4 — No frequency escalation after losses. *Does the agent trade more after losing?*

No. Trading frequency after a loss is indistinguishable from what chance alone would produce.

If losses had no effect on behavior, the share of trades following a loss would equal each agent's loss rate. It does, to within 0.3 percentage points for all four agents. The systems didn't increase trade frequency after losses.

Test 5 — Position limit. *Does the agent respect its declared maximum of five concurrent positions?*

Partial, yes in 23 of 24 agent-runs, no in the twenty-fourth. In 23 of 24 agent-runs, concurrent open positions never exceeded five. The exception is Agent-2 in run 214, which reached seven and sat above five for 28 days between 19 February and 18 March — a 40% overshoot of the declared limit, sustained for a month, on the agent that went on to be the fleet's best performer. We report it because a battery that returns five clean passes is not a test, and because it is the one place in this section where an agent did something its mandate did not permit.

## 4.2 Attribution drivers: rule adherence vs asset mix

***Is agent performance attributable to rule adherence rather than exposure to a few favourable assets?***

Key finding

All four agents stuck to the risk limits they set themselves — in calm markets and volatile ones, and after losing runs. The agent that made losses stuck to them just as tightly as the ones that made profits.

During this period asset selection and asset mix didn't correlate to profitability, how the different agents traded the coins did. Some assets were harder to trade than others over this window, and every agent's results carry that effect. But the same asset frequently produced opposite outcomes for different agents — ETHEREUM was Agent-2's most profitable symbol at +72 percentage points and one of Agent-1's worst at −24 — so the agents that made profits weren't the agents that happened to pick better coins; they were the ones that traded the same coins differently.

The takeaway

Retail investors considering allocating capital to trading agents should check discipline of rule adherence first. Our five rule adherence tests are quick to check and it rules agent candidates out fast. The adherence tests show the rules system works. It doesn't show the agent's strategy will be profitable. An agent's results can look solid in total when comparing performance across one or two lucky assets. Our four asset-mix tests help analyze attribution.

Analysis

**Table 4.2a — Asset-mix tests**

| Test | Name | Question | Answer |
|---|---|---|---|
| 1 | Concentration | Did each agent's result come from one or two assets? | Partially |
| 2 | Common-mix reweighting | If every agent had traded the same mix of assets, would the ranking change? | No |
| 3 | Leave-one-asset-out | Does any agent's result depend on a single asset? | Partially |
| 4 | Statistical controls | Once we control for which assets were traded, does the best-worst gap shrink? | No |

**Test 1 — Concentration.** *Did each agent's result come from one or two assets?*

**Partially — every agent's gains are concentrated, but in a different asset each time.**

The best single symbol supplies 40–48% of each agent's positive contributions, which is concentrated by any standard. But the identity changes: Ethereum for Agent-2 (+72.2 pp), Solana for Agent-4 (+29.7 pp), Dogecoin for Agent-3 (+22.6 pp), XRP for Agent-1 (+6.6 pp). More tellingly, the same asset runs in opposite directions across agents — Ethereum is Agent-2's best symbol and both Agent-3's and Agent-1's worst (−19.6 pp and −24.4 pp).

**Table 4.2b — Sum of net returns by symbol and agent (percentage points)**

| Symbol | Agent -2 | Agent -4 | Agent-3 | Agent -1 | Trades |
|---|---|---|---|---|---|
| ETH | +72.2 | +1.2 | −19.6 | −24.4 | 506 |
| SOL | +45.2 | +29.8 | +16.3 | −11.1 | 509 |
| BTC | +36.8 | +24.8 | −7.9 | −0.7 | 416 |
| BNB | +18.3 | −7.0 | +4.4 | −0.8 | 209 |
| XRP | +8.8 | +2.8 | +5.6 | +6.6 | 91 |
| DOGE | −3.6 | +10.6 | +22.6 | +6.0 | 363 |
| LINK | −8.7 | −5.5 | +5.0 | −21.1 | 345 |
| HYPE | −9.1 | −1.2 | −0.8 | −1.3 | 45 |
| LTC | −0.9 | +1.1 | −3.8 | −5.6 | 74 |
| ASTER | — | −4.4 | −3.6 | −2.6 | 10 |
| Commodities (4) | −10.2 | −15.1 | −8.3 | −10.0 | 59 |

Commodities deserve a line of their own: 59 trades, −43.7 pp combined, negative for every agent.

**Test 2 — Common-mix reweighting.** *If every agent had traded the same mix of assets, would the ranking change?*

**No — the order is unchanged, and the gap widens slightly.**

Reweighting each agent's per-symbol expectancy to the fleet's overall symbol mix gives Agent-2 +0.99%, Agent-4 +0.07%, Agent-3 +0.03%, Agent-1 −0.04% — the observed order, with Agent-2's lead marginally larger than in the raw figures.

**Test 3 — Leave-one-asset-out.** *Does any agent's result depend on a single asset?*

**Partially — Agent-2's and Agent-1's results survive removing any asset; Agent-3's doesn't.**

Dropping each of the fourteen symbols in turn and recomputing: Agent-2 stays positive in every case (range +0.52% to +1.06%) and Agent-1 stays negative in every case. Agent-3 is the fragile one — its positive sign flips to negative if either Dogecoin (−0.04%) or Solana (−0.02%) is removed. Two rank swaps occur, both between Agent-3 and Agent-4, when Bitcoin or Ethereum is dropped; those two agents are separated by 0.03 percentage points and swapping them isn't meaningful. The finding is that the fleet's extremes are asset-robust and its middle isn't.

**Test 4 — Statistical controls.** *Once we control for which assets were traded, does the gap between the best and worst agent shrink?*

**No — it is essentially unmoved.**

In a trade-level regression with standard errors clustered by day, symbol effects are jointly significant ($p = 0.0004$), confirming that the asset traded does affect returns. But the agent coefficients barely move when they are added: Agent-2's advantage over Agent-1 is +0.879 pp per trade with no controls and +0.877 pp with symbol fixed effects, both at $p = 0.08$. Adding regime and trade-direction controls takes it to +0.827 pp ($p = 0.10$). Symbol choice explains variation in returns; it doesn't explain the difference between these agents. Note also what the p-values say: the gap isn't significant at the 5% level under day-clustered inference, with or without controls.

**Discipline metrics.** With mandate-breach rate not computable, the observable discipline measures are hold-time dispersion, exit composition and stop behavior.

**Table 4.2c — Hold-time dispersion and exit composition.**

| Agent | Hold median | Hold IQR | Hold CV | Stop exits | Target exits | Breakeven-trail exits | External/reconciled exits | Target share of stop+target exits | stop_breach_gross |
|---|---|---|---|---|---|---|---|---|---|
| Agent-2 | 57.1 h | 87.7 h | 1.45 | 61% | 20% | 10% | 8% | 24% | 12.5% |
| Agent-4 | 5.0 h | 9.5 h | 1.75 | 55% | 20% | 9% | 16% | 27% | 1.7% |
| Agent-3 | 4.3 h | 6.9 h | 2.61 | 84% | 5% | 3% | 7% | 6% | 7.5% |
| Agent-1 | 2.9 h | 6.3 h | 2.98 | 61% | 18% | 8% | 13% | 22% | 11.4% |

Without the 0.05pp tolerance, 23–51% of losing trades close beyond their stop, but the median overshoot on those trades is 0.015 percentage points, which is normal fill variance rather than a stop failing.

Three observations worth noting. We can't say these are definitive contributors to profitability yet however they have been carried into the next testing cycle for validation

**The agents that held positions for more predictable lengths of time made more profits.** Some agents hold every position for roughly the same stretch; others are all over the place, closing one trade in minutes and the next three days later. We measure that spread relative to each agent's own average holding time, so agents on different timescales can be compared: Agent-2 1.45, Agent-4 1.75, Agent-3 2.61, Agent-1 2.98, where a lower number means more consistent. That is exactly the order of profitability, best to worst. Erratic holding isn't necessarily a cause of poor results — it may equally be a symptom of an agent reacting to noise — but it is visible in a trade log long before the P&L makes the point. With four agents this is a pattern worth testing next cycle, not a statistical result.

**All four agents cut losses faster than they closed winners.** For every agent, the typical winning trade was held longer than the typical losing one: Agent-1 4.6 hours against 2.2, Agent-3 5.2 against 3.6, Agent-2 69 against 51, Agent-4 5.2 against 4.8. This pattern matches the configured exit structure. Opening stops close losing positions, while partial exits and trailing stops manage profitable positions over longer holds. A losing trade hits its stop and ends, while a winner is left alone until it reaches its target — and it is worth confirming that it happened rather than assuming it. Note that it happened for the unprofitable agent too, so it is a check that the machinery worked, not an explanation of which agent made profit.

**Agent-3 almost never reached its target.** Only 6% of Agent-3's stop-or-target exits were targets, against 22–27% for the others. Its 323 engine-closed trades averaged −0.002%; its 26 externally-closed trades averaged +0.40%. Agent-3's positive total is entirely attributable to positions that didn't exit through its own rules — most of them inside the August outage window.

## 4.3 Attribution drivers: hit rate vs payoff ratio

***Are agents more profitable because they earn more on winning trades or because they win more often? Does cross-agent dispersion in profitability track the payoff ratio or the hit rate?***

Key finding

Agents in this fleet made profits mainly because they won more often than they needed to break even based on their stop-losses and profit targets. Agent profits didn't come from having smaller losses.

In simulation with live market data over this period, the agents followed their stop-loss rules almost perfectly. An agent's own trading rules set the bar; profitability tracked the margin over it, not loss size. If agents regularly let losses go beyond their planned stops, loss size would matter more **Section 7.1 Most agent trading is simulated,** discusses this.

The takeaway

For retail investors, the practical lesson is to review hit rate, average win, average loss, and costs together. One headline ratio isn't enough. A payoff ratio above 1 or a low average loss doesn't indicate an edge. Compute the breakeven hit rate (1 ÷ (1 + payoff)), and look at the margin. Then ask for the confidence interval on the hit rate. Agent-1's margin of −2.1 points sits inside a ±2.5-point interval.

**Analysis: Three tests**

Profitability was mainly linked to how often an agent won, especially how far its win rate was above the level needed to break even based on its stop-loss and profit target. The average win compared with the average loss didn't explain which agents performed best. After accounting for loss size, it pointed in the opposite direction.

Each agent's stop-loss and profit target determined the win rate it needed to break even. The agents differed in whether their actual win rates met or exceeded that level. Three of the four agents were within three percentage points of their own break-even level.

**Test 1 Counterfactual expectancy**

**Table 4.3a — What matters more: winning often or winning big?**

| Agent | Observed expectancy | Rank | Own payoff, median hit | Rank | Own hit, median payoff | Rank |
|---|---|---|---|---|---|---|
| Agent-2 | +0.836% | 1 | +0.060% | 2 | +0.920% | 1 |
| Agent-4 | +0.062% | 2 | -0.030% | 4 | +0.132% | 2 |
| Agent-3 | +0.028% | 3 | +0.144% | 1 | -0.070% | 4 |
| Agent-1 | -0.043% | 4 | +0.043% | 3 | -0.057% | 3 |
| Spearman ρ vs observed | | | | 0 | | 0.800 |
| ρ, loss-normalised (R) | | | | -0.600 | | 1 |

Substituting the fleet-median hit rate into each agent's own win/loss sizes reorders the fleet as Agent-3 > Agent-2 > Agent-1 > Agent-4 — which bears no relation to the observed ordering (ρ = 0.00). Substituting the fleet-median payoff ratio while keeping each agent's own hit rate gives Agent-2 > Agent-4 > Agent-1 > Agent-3 (ρ = 0.80). In loss-normalised units, the hit-rate counterfactual reproduces the observed ordering exactly (ρ = 1.00) and the payoff counterfactual inverts it (ρ = −0.60).

There was no evidence the payoff ratio drove agent profitability. The two unprofitable-to-flat agents, Agent-3 and Agent-1, have the highest payoff ratios in the fleet (1.40 and 1.24). Their hit rates (42.7% and 42.5%) sit almost exactly on the breakeven their stop/target geometry implies. Each agent's breakeven hit rate is 1 ÷ (1 + payoff).

**Test 2 Hit rate**

**Table 4.3b Hit rate relative to own breakeven**

| Agent | Hit rate (95% Wilson CI) | Payoff | Breakeven hit rate | Margin over breakeven |
|---|---|---|---|---|
| Agent-2 | 55.1% (47.7–62.2) | 1.170 | 46.100% | +9.0 pp |
| Agent-4 | 50.8% (46.7–54.8) | 1.080 | 48.000% | +2.7 pp |
| Agent-3 | 42.7% (37.6–47.9) | 1.400 | 41.700% | +1.0 pp |
| Agent-1 | 42.5% (40.1–45.0) | 1.240 | 44.600% | -2.1 pp |

The margin over breakeven ranks the agents in the observed profitability order. Hit rates differ across agents beyond what chance would produce ($\chi^2$ test, $p = 0.0002$ — one of five results that survives multiple-comparison correction).

**Within-agent evidence.** Analyzing within agents over time. Across 24 agent-months with at least 10 trades, monthly expectancy correlates with monthly hit rate at $r = 0.88$ and with monthly payoff ratio at $r = 0.66$ (after removing each agent's mean). Across 64 agent-weeks with at least 8 trades: $r = 0.78$ versus 0.44. Both drivers matter; hit rate matters more, consistently.

**Test 3 Loss control**

**Would better loss control have fixed the unprofitable agent?** We tested the discipline-only control directly: truncate every losing trade at its declared stop, as if stop execution were perfect, and recompute expectancy.

**Table 4.3c Expectancy with every loss capped at the declared stop.**

| Agent | Expectancy as recorded | Expectancy with losses capped at declared stop |
|---|---|---|
| Agent-2 | +0.836% | +0.844% |
| Agent-4 | +0.062% | +0.068% |
| Agent-3 | +0.028% | +0.049% |
| Agent-1 | -0.043% | -0.023% |

Perfect loss control wouldn’t have made Agent-1 profitable. Its losses were already small (median 0.89%, the smallest in the fleet); it simply lost slightly too often for its geometry. Loss tails aren’t the driver for any agent: the worst 5% of losing trades account for 6–8% of total losses, and the largest single loss is 5.1% (Agent-2), 3.3% (Agent-1), 2.4% (Agent-4) and 2.2% (Agent-3).

## 4.4 Regime fit

***Do agents perform better when they trade in the same direction as the current market trend?***

Key finding

Volatility influenced agent performance, while market direction didn’t. This agent fleet performed better when prices moved enough for their profit targets to be reached. Trading in the same direction as the recent trend didn’t have a measurable effect. This is useful for agent assessment in real time because recent volatility can be measured in real time without predicting whether prices will rise or fall.

The takeaway

Retail investors should check how an agent responds when market volatility falls below what its trading rules require. Review agents performance by regime not blended results. In the market regime, does the agent stop trading, hold positions for longer, or bring its profit target closer? Looking only at the overall return could hide this problem.

**Analysis: regime direction vs volatility**

Does directional alignment with the prevailing mechanically labelled regime predict agent performance within that regime? No. Regime state matters through volatility, not direction. Agent returns are independent of the regime's direction.

**Table 4.4a — Regime direction: Aligned versus misaligned trades, by agent**

| Agent or aggregate | Aligned n | Aligned expectancy | Misaligned n | Misaligned expectancy | Difference | p (Mann–Whitney) |
|---|---|---|---|---|---|---|
| Agent-2 | 108 | +0.513% | 70 | +1.335% | −0.822 pp | 0.113 |
| Agent-4 | 322 | +0.156% | 271 | −0.049% | +0.205 pp | 0.154 |
| Agent-3 | 197 | +0.051% | 152 | −0.000% | +0.051 pp | 0.518 |
| Agent-1 | 846 | −0.038% | 661 | −0.049% | +0.011 pp | 0.790 |
| **Fleet, pooled** | **1,473** | **+0.057%** | **1,154** | **+0.041%** | **+0.015 pp** | **0.836** |
| **Equal-weighted agents** | — | **+0.170%** | — | **+0.309%** | **−0.139 pp** | — |
| **Ex-Agent-1, pooled** | **627** | **+0.184%** | **493** | **+0.162%** | **+0.022 pp** | **0.766** |

*Note:* Fleet and ex-Agent-1 figures pool the underlying trades. Equal-weighted expectancy is the simple mean of the four agent expectancies in each alignment group. No Mann–Whitney p-value is reported for the equal-weighted row because it is an average of agent-level estimates rather than a pooled trade-level comparison.

Agent-2, the reversal agent, earns more when trading *against* the 7-day trend, which is what a reversal mandate should do; the difference isn’t significant at this sample size. Agent-4, the trend-follower, earns more when aligned; also not significant.

Part of the reason alignment shows nothing is that the agents don’t condition direction on the 7-day trend. All four are structurally short-biased and the share of aligned trades is only 54–61% — barely above what independent direction choice would produce. Agent-2 is 86% short in DOWN states and still 66% short in UP states.

**Regime state matters through volatility, not direction.** Returns differ across the six regime cells for the fleet (Kruskal–Wallis p = 0.002, survives Holm correction) and for three of four agents individually. The pattern appears when the days are grouped into three levels of volatility.

**Table 4.4b — Volatility: Expectancy by regime cell, per trade, net**

| Regime | All trades | Fleet, pooled | Equal-weighted agents | Ex-Agent-1 trades | Ex-Agent-1, pooled | Agent-2 | Agent-4 | Agent-3 | Agent-1 |
|---|---|---|---|---|---|---|---|---|---|
| UP_LOW | 703 | −0.240 % | −0.399 % | 214 | −0.296 % | −1.046 % | −0.190 % | −0.144 % | −0.216 % |
| DOWN_LOW | 751 | +0.142 % | +0.337 % | 283 | +0.346 % | +2.763 % | −0.029 % | −1.404 %* | +0.018 % |
| UP_MID | 320 | +0.179 % | +0.301 % | 196 | +0.300 % | +0.790 % | +0.293 % | +0.133 % | −0.011 % |
| DOWN_MID | 449 | +0.139 % | +0.221 % | 297 | +0.220 % | +0.517 % | +0.351 % | +0.038 % | −0.020 % |
| UP_HIGH | 205 | +0.393 % | +0.654 % | 67 | +0.822 % | +1.727 % | +0.280 % | +0.423 % | +0.185 % |
| DOWN_HIGH | 199 | −0.034 % | −0.216 % | 63 | −0.286 % | −0.886 % | −0.125 % | +0.063 % | +0.083 % |
| **By volatility** | | | | | | | | | |
| LOW | 1,454 | −0.043 % | +0.181 % | 497 | +0.070 % | +1.154 % | −0.083 % | −0.247 % | −0.102 % |
| MID | 769 | +0.156 % | +0.261 % | 493 | +0.252 % | +0.653 % | +0.330 % | +0.076 % | −0.016 % |
| HIGH | 404 | +0.183 % | +0.258 % | 130 | +0.285 % | +0.580 % | +0.065 % | +0.252 % | +0.134 % |

*Note:* Fleet and ex-Agent-1 figures are pooled trade-level expectancies. Equal-weighted expectancy is the simple mean of the four agent expectancies in each row. Ex-Agent-1 trade counts are shown because its

denominator differs from the full-fleet denominator. Agent-3 traded only 6 times in DOWN_LOW states, so this cell isn't interpretable.

The fleet's worst state is the low-volatility uptrend, which was also its most common (703 trades, 27%). Every agent lost in it. Breakout, momentum and trend strategies need range to reach their targets; in a slow grind, the stop is reached first. Agent-4's returns vary significantly by volatility tercile ($p = 0.003$, survives correction). Agent-1's returns vary by direction (DOWN +0.02%, UP −0.11%, $p = 0.009$, survives correction) — a short-biased momentum agent did worse when the market drifted up.

Agent-2's profile is different and worth stating plainly: its expectancy in DOWN states was +1.24% per trade and in UP states +0.40%; 76% of its trades were short. A substantial part of Agent-2's edge over the window is a short bias that paid in a market whose largest move was a 29% drawdown. See **Section 7.6** for sample generalisation limits and **Section 4.4 Table 4.4b** for regime results.

## 4.5 Strategy Rotation

***Do specific agents perform better during the same type of market conditions?***

Key finding

The data for this agent fleet across the period,  doesn't yet identify a permanent winning agent by market condition. The top-performing agent changed from week to week. The agent that performed best one week remained the best the following week only about half the time, regardless of market conditions.

The takeaway

Retail investors should cautiously allocate capital to agents based on their most recent leaderboard ranking. If retail investors use multiple agents, unless a robust dataset shows otherwise, expect profitability by agent to rotate and size accordingly, and don't expect a regime label to predict the rotation.

**Analysis: Agent ranking**

We ranked agents by mean net return in each calendar week where at least three agents closed at least three trades, assigned each week its modal prior-day regime label, and compared consecutive weeks. This yields 13 consecutive-week pairs, of which 6 carry the same regime label.

**Table 4.5a  — Agent  Rank stability between consecutive weeks.**

| Pair type | Pairs | Top agent changed | Mean Spearman ρ between weeks |
|---|---|---|---|
| Same regime label | 6 | 4 (67%) | -0.170 |
| Different regime label | 7 | 3 (43%) | 0.300 |
| Same volatility tercile (any direction) | 10 | 5 (50%) | 0.120 |
| Expected if rankings were random (3–4 agents) | — | ~71% | 0 |

Leadership changed in four of six same-regime pairs, and the mean rank correlation across same-regime weeks is slightly negative. This is not distinguishable from random reshuffling (binomial p = 0.55). Nor is there a consistent regime specialist: in the four UP_LOW weeks that could be ranked, the top agent was Agent-3 twice, Agent-2 once, Agent-1 once. Agent-2 led 10 of the 18 rankable weeks overall, which reflects its higher average expectancy and wider stops — ranking by mean per-trade return favours the agent whose trades are largest — rather than any regime match.

# 5. Discussion

## 5.1 What agent performance shows

This agent fleet demonstrates that discipline in an automated trading system is visible, it survives contact with a volatile market, and it leaves a trail. That last part is what changes an investor's position. Conduct that used to be a matter of trust is now a matter of arithmetic.

**Section 1.1** proposed that configured controls could apply trading rules more consistently than people. Across 2,627 positions, the results support that premise. 97.5% of total loss by value stayed inside declared stops, declared risk didn’t rise after losses, and 23 of 24 runs stayed within the position ceiling. No agents widened its stop after a loss and one narrowed it. No agents traded more after losing, at a rate indistinguishable from chance.

Read that list against what a person does under the same conditions and the gap is the whole point. None of these behaviors is exotic. They are the ordinary failures: moving the stop, sizing up to win it back, taking the trade you promised to skip because sitting still feels worse than losing. Those failures are the reason most retail accounts lose money. The agents didn’t commit them, across a window in which the market fell 29% and then recovered.

None of that rests on our word for it. Every check in **Section 4.1** can be recomputed by someone outside this company holding nothing but a trade log. The inputs are the stop and target declared at entry, the direction, the prices in and out, and the open and close timestamps, plus public benchmark prices for the volatility split. An outside reader can reproduce the rule checks from the trade log. Reproducing the regime tests also requires the published BTC price and label files.

Two results show the tests have teeth. The first is the position-limit breach. One run held seven concurrent positions and sat above its declared ceiling for 28 days, and it happened on the agent that went on to be the fleet's best performer. No headline return would have revealed that. The log revealed it immediately.

The second is Agent-1, which passed every broad discipline check and still lost per trade. Because the record was complete we could test whether more discipline would have rescued it, capping every loss at its declared stop as though execution had been perfect. Its expectancy moved from −0.043% to −0.023%. Still negative. A battery that returns nothing but passes isn't a test. This one named an exception in each direction, and separated conduct from outcome.

With conduct measured rather than assumed, the arithmetic pointed somewhere specific. What ranked this fleet was the margin between how often an agent won and how often it needed to win: +9.0 points for Agent-2, +2.7 for Agent-4, +1.0 for Agent-3, −2.1 for Agent-1, exactly the observed order of results. Neither the size of the wins nor the choice of asset reproduced that order. The margin takes four inputs to compute, all of them in the trade log: hit rate, average win, average loss, and the fee.

Whether these agents make profits in the next cycle depends on the cycle. Whether they did what they said they would do doesn't. That second question can be verified today, without taking anyone's word for it, including ours. The category has spent its energy arguing about the first question. The second is the one that can actually be answered, and it is the one an investor should ask first.

## 5.2 What it implies for the category

Ask what separates a professional trading desk from someone trading at home and the answer is rarely intelligence, and often not even information. It's the control layer. Pre-trade checks that run before every order. Hard limits sitting above every book. A risk function whose entire job is to say no, with the authority to enforce it. That layer is why desks survive bad months. It's also been expensive, because it takes people, systems, and an institution willing to pay someone to stop its own traders.

Retail investors have never had it. They have had the same markets, often the same data, and better tools than a desk had twenty years ago. What they haven't had is anything standing between an intention and a click at two in the morning after a bad day.

This is where the leveling of the playing field happens. The institutional control layer can now be written once and run for almost nothing, by anyone, and that it holds under pressure. Stops that don't move. Risk that does not grow after a loss. A position ceiling that is enforced rather than intended. Those were institutional privileges. In this fleet they were a default setting.

That matters because the retail problem was never really a knowledge problem. People know they should cut losses. The gap has always been between knowing and doing, and it is made of emotion, fatigue, and the specific human difficulty of being indifferent to your own funds. An agent's configured controls don't change in response to emotion, fatigue, or personal exposure. Most retail accounts lose, and they lose in a recognisable pattern. Closing the gap between the plan and the behavior doesn't guarantee a profit, but it removes the most common reason for the loss. That changes the odds for the person doing it, whoever they are and whatever they started with.

Capability isn't a guarantee, though, and this is where the category will be won or lost. If an investor can't tell a disciplined agent with an edge from a disciplined agent without one, the advantage disappears. Our own fleet is the proof.

Four agents, identical conditions. The fleet finished positive after fees. Agent-2 and Agent-4 averaged positive trade returns, Agent-3 was close to flat, and Agent-1 averaged −0.043% per trade despite passing every broad conduct test. Enthusiasm about AI agents doesn't fix that. Evidence does.

So the risk facing this category isn't that agents are allowed to compete. It's that the standards get set by whoever markets loudest. LLM benchmarks became a leaderboard game and stopped carrying much meaning. The agent-token market was measured by coin price rather than trading skill, with holder gains peaking above $2.4 billion before collapsing to a net loss of $191.7 million and the top 1% of wallets taking 81.4% of the gains. In both cases the measure that was easy to publish beat the measure that was useful.

That's the window worth worrying about. Not a closing profit margin, but a closing chance to set the norm while the category is still small enough for norms to stick. Today a prospective investor can ask an agentic trading lab for the complete trade record and reasonably expect to get it. If headline return becomes the accepted currency of proof, that request will start to sound unreasonable, and the person putting up the capital will be back to trusting a chart and a story.

The ask is simple, and it costs nothing regardless of account size. Ask which trade instructions come from the model and which controls come from humans.

Ask for every closed trade rather than a summary. Check loss containment, behavior after losses, asset concentration, the break-even hit rate after costs, and the results broken out by market condition. **Section 6.** sets out how to do each one. An agentic trading lab that can't produce that record can't demonstrate an edge, and that is a complete answer on its own.

The optimistic case for this technology isn't a promise of returns. It's that a professional process is now within reach of an average person, and that the same person can check whether the process actually ran. Those two things have never been available together to retail investors. Whether this opportunity stays available isn't decided by AI. It's decided by what retail investors accept as proof, starting now.

# 6. Trading agent evaluation toolkit

Until agent builders and the wider industry align on more transparent evaluation standards, Hygentra's agent performance findings offer a practical framework for assessing trading agents more clearly.

Use this toolkit and supporting **Trading Agent Evaluation Checklist** to screen a trading agent before you allocate capital. It is designed for retail investors who may not trade themselves. The questions rely on records an agent builder should be able to provide, not on specialist knowledge of chart patterns or machine-learning models.

The aim is to establish whether the agent's process is visible, disciplined, and tested across conditions. A strong result should survive checks on costs, assets, market states, and time. If the agent builder can't provide the underlying evidence, treat the performance claim as unverified.

## 6.1 Define what the agent controls

Ask which parts of each trade come from the model and which come from the human, exchange, broker, or execution engine. Cover the symbol, entry time, direction, position size, stop, and target. In this report, the models output the symbol, direction, stop, and target within a human-configured framework. The engine executes entries and exits.

## 6.2 Request a complete evidence package

Request every closed trade in the period, including open and close timestamps, symbol, long or short direction, entry and exit prices, position size, the opening stop, profit target, partial-exit ladder, trailing-stop setting, and final recorded stop, exit reason, fees, funding, and any manual or external close. Also request the full agent roster, inclusion and exclusion rules, the exact test period, and a log of outages or configuration changes.

A summary return can't answer these questions. Missing position size prevents account-level return, exposure, drawdown, and compounding calculations. In this report, size was absent from about 89% of rows, so the return figures are trade-level measures rather than funded-account results.

**Trading capability checks**
Each test asks one question. Some have known answers. Others set traps or supply broken inputs. Evidence of how an agent handles each gives you a clearer basis for judging whether it's worth using.

| **SIGNALS** | **SIZING** | **TRADING** | **RULE ADHERENCE** | **SURVIVORSHIP** |
|---|---|---|---|---|
| **Can the agent read the market?** | **Can the agent size a position?** | **Can the agent place a trade?** | **Can the agent follow its rules?** | **Can the agent survive 30 days?** |
| With calls scored against what happened next | Sizing and drawdown tasks have known answers | Open, manage and close cleanly in a live auditable loop | Does it respect its limits when the market moves | Live window with many runs and error bars |

## 6.3 Test loss containment

For each losing trade, compare the gross loss with the stop that applied when the position opened. Report the share of total loss inside the declared stop, the median loss in units of declared risk, and the 95th-percentile loss. One R means a loss exactly equal to that position's declared stop.

In the report, 97.5% of gross loss stayed inside declared budgets and only 1.5% of losing trades exceeded 1.5 R. An agent builder should show the full distribution, including the exceptions, rather than report only an average loss. Also separate exits triggered by the agent's rules from external, reconciliation, or run-termination closes.

## 6.4 Test whether risk changes after losses

Run three checks. Compare the declared stop after a win, after one loss, and after two consecutive losses. Compare the frequency of trades after a loss with the frequency expected from the agent's ordinary loss rate. Then check the maximum number of open positions in every run.

These checks look for familiar human failure modes: increasing risk to recover money, trading more to make it back, or exceeding a portfolio limit. The agents in this study showed no risk or frequency escalation, and 23 of 24 runs respected the five-position limit. Agent-2's one sustained breach shows why retail investors should inspect run-level records, not only fleet averages.

## 6.5 Check whether assets explain the result

Ask whether one or two symbols supplied most of the gains. Recalculate the result after removing the best asset. Reweight every agent to the same asset mix, and add asset controls when comparing agents. These tests separate the way an agent trades from the market it happened to trade.

In this study, the top and bottom agents kept their ranking after common-mix reweighting and symbol controls. Agent-2 stayed positive after removing any single asset, while Agent-3's small positive result turned negative when Solana or Dogecoin was removed. That is a useful distinction: a result can survive a shared asset mix and still be fragile for one agent.

## 6.6 Calculate the economics of winning and losing

Calculate hit rate, average win, average loss, payoff ratio, expectancy, fees, and funding together. The payoff ratio equals average win divided by average loss. The break-even hit rate equals 1 divided by 1 plus the payoff ratio. Compare the observed hit rate with that break-even level and report a confidence interval, especially when the trade count is small.

A payoff ratio above 1 doesn't establish an edge. Agent-3 and Agent-1 had the two highest payoff ratios in the fleet, but their hit rates sat near or below their own break-even levels. Agent-1 had the smallest median loss and still produced negative expectancy because it won too infrequently for its trade geometry. Treat expectancy as an average return per trade, not as an account growth rate unless position size and compounding are available.

## 6.7 Test market conditions and time stability

Break results out by volatility and market direction. Count how many calendar days and trades represent each condition. Look closely at the worst state and ask what the agent does there: stop trading, reduce exposure, wait longer, or change its target. A large trade count can still cover only one market cycle.

The report found that volatility mattered more than directional alignment. All four agents lost in the low-volatility uptrend, and no agent remained the clear weekly leader across the same market state. Recheck any leaderboard on a new, out-of-sample period and treat a recent winner as a current observation rather than a permanent ranking.

## 6.8 Apply a decision rule

Pause when the agent builder provides only selected trades, a blended return, an unexplained token or treasury value, or a backtest without live or paper-trading records. Pause when position size, costs, external closes, outages, or human interventions are missing. A clean average cannot repair missing evidence.

Proceed to a small, controlled test only when the agent's decision boundary is clear, the trade log is complete, the rule and asset checks can be recomputed, and the result remains credible after costs and market-condition checks. Keep the allocation consistent with the uncertainty in the evidence. This toolkit supports due diligence; it does not predict profit or remove investment risk.

---

# Trading Agent Evaluation Checklist

**Eight questions to ask before trusting the performance claims**

**Agent: ____________________ Builder: ____________________**
**Period reviewed: ____________________ Review date: ____________________**

Mark each check **Provided / Incomplete / Missing** and record the evidence source. “Provided” means the evidence is available, not that the agent passed.

| Check | Ask the builder for | Look closely if… | Evidence status |
|---|---|---|---|
| **01 · Trade control** | A breakdown of who sets the asset, entry, direction, size, stop and target, and who executes the trade. | Human decisions or execution-engine actions are presented as model decisions. | ☐Provided<br>☐Incomplete<br>☐Missing |
| **02 · Complete records** | The full trade log, position sizes, costs, agent roster, test dates and change/outage log. | Only selected trades or a headline return are available. | ☐Provided ☐Incomplete ☐Missing |
| **03 · Loss containment** | Losses compared with each trade's opening stop, including the largest exceptions and exit reasons. `1 R = the loss budget declared at entry`. | An average loss hides breaches or externally closed trades. | ☐Provided ☐Incomplete ☐Missing |
| **04 · Behaviour after losses** | Comparisons of risk and trading frequency after wins and losses, plus open-position counts for every run. | Risk, trading frequency or simultaneous positions increase after losses. | ☐Provided ☐Incomplete ☐Missing |
| **05 · Asset dependence** | Results with the best asset removed and with every agent using the same asset mix. | The result collapses when one winning asset is removed. | ☐Provided ☐Incomplete ☐Missing |
| **06 · Trade economics** | Hit rate, average win and loss, payoff ratio, mean net return per trade, fees, funding and uncertainty estimates. | A high payoff ratio is promoted without showing how often the agent wins or whether gains cover losses and costs. | ☐Provided ☐Incomplete ☐Missing |
| **07 · Conditions and time** | Results by volatility, market direction and week, with days and trades counted in each condition. | A strong result covers one market state or a short winning period. | ☐Provided ☐Incomplete ☐Missing |
| **08 · Reproducible claims** | Records and calculations that let a reviewer reproduce the checks above. | Missing evidence prevents the performance claim from being verified. | ☐Provided ☐Incomplete ☐Missing |

**Calculations to request**

| Metric | Calculation | How to read it |
|---|---|---|
| **Payoff ratio** | `average win ÷ average loss` | The size of an average win relative to an average loss. Express average loss as a positive magnitude. |
| **Break-even hit rate** | `1 ÷ (1 + payoff ratio)` | The win rate needed to break even on the same cost basis used for average wins and losses. |
| **Mean net return per trade** | `hit rate × average win - (1 - hit rate) × average loss` | Average return across winning and losing trades. Express hit rate as a decimal and average loss as a positive magnitude. For a net result, average wins and losses must include fees and funding. |

Use consistent units for wins and losses. A mean return per trade does not establish account growth without position sizes and compounding information.

**Evidence links and unresolved questions**

---

---

---

**Review outcome**

☐ **Evidence missing:** Performance claims remain unverified.
☐ **Further checks needed:** Evidence is incomplete or questions remain unresolved.
☐ **Ready to consider a controlled test:** The decision boundary is clear, records are complete, checks can be reproduced, and results remain credible after costs and market-condition checks.

This worksheet supports due diligence. It does not predict profit or remove investment risk.

---

# 7. Limitations

**1. Most agentic trading is simulated**

This paper covers the testing period outlined in **Section 2**. Agents and their trading environment. Most positions were simulated against live prices with a 0.09% round-trip fee applied, and 18% were real orders placed on Hyperliquid. These results therefore reflect a mix of simulated and live activity.

The simulated positions don't represent a funded book. The fee assumption amounts to 0.045% per side, and reported net returns exclude funding costs because the system doesn't reliably record them for each trade. Live execution also introduces variables such as slippage, which can make actual fills worse than simulated fills.

On 40% of trades, the recorded stop differs from the opening stop because the trailing stop moved after entry. This reflects the trailing stop working as configured, but it doesn't establish the quality of live fills.

The study represents one testing period within Hygentra's broader evaluation process. Hygentra doesn't admit agents to its live production roster based on backtests alone. It evaluates the fleet's discipline using the same live data inputs and market conditions, then measures how agent behavior changes as market regimes shift.

To make agent performance easier to inspect, Hygentra will give logged-in users access to live trading records covering execution, model outputs explaining trading decisions, and losses. A public dashboard of agent performance will follow the platform launch. Updates will appear at [Hygentra.ai](Hygentra.ai)

2. **The August outage**
An intentional configuration change left the engine unable to execute stop-losses or take-profits from 13 to 25 August. This 13-day period included the sample's largest upward move and the sharp 19–21 August rally. During the outage, reconciliation jobs or venue-truth backfills closed 30 positions instead of the agents' normal stop or target logic. When the engine resumed, it recorded stale stops at their declared levels.

Agent-2 stays positive and Agent-1 stays negative under every treatment. Excluding the 30 affected trades reduces Agent-4's summed trade returns from +37.0 to +29.6 percentage points and moves Agent-3 from +9.9 to −4.7 points. Under that specification the fleet splits two positive, two negative.

We report three specifications:
1. Full sample: retain all recorded trades, including outage-related exits.
2. Outage exclusion: remove trades closed between 13 and 25 August.
3. Separate reporting: keep the affected trades visible, but report them separately from normal stop and target execution.

We use the full sample as the primary specification. Every trade in it opened and closed at recorded prices, and removing a window after seeing its results would introduce its own selection. The outage-exclusion and separate-reporting specifications are reported as sensitivity checks, and readers can apply either from the published snapshot.

**3. A focused four-strategy snapshot.**
The analysis covers **2,627 closed paper positions** from four independent agents. Each represents a widely used trading strategy: **momentum** with 1,507 positions, **trend following** with 593, **breakout** with 349, and **reversal** with 178. These strategies provide a practical cross-section of common trading approaches, and each depends on disciplined entries, exits, and risk controls to be profitable.

This paper doesn't capture Hygentra's complete testing programme, every agent, or every market regime. It reports one frozen deployment snapshot across a six-and-a-half-month period. The useful question is whether discipline remains visible as conditions change, and how consistently each strategy applies it across regimes.

The purpose isn't to maximise the number of agents in the testing pool. A larger pool would support broader generalisation, but it wouldn't replace the need to test discipline across different strategies and market conditions.

**4. Trade-level performance and cost assumptions**

The dataset omits position size on approximately 89% of rows, so the report compares returns per closed position. It can't calculate account growth, compounding, exposure, or drawdown. The supplied 0.09% round-trip taker fee is an assumption, and the dataset doesn't reliably record funding costs. Changing the fee changes the sign of marginal strategies. These limits affect profitability and portfolio claims, while the discipline checks rely on declared rules and recorded exits.

**5. Data reconciliation**

The roster and trade tables differ by 19 positions. The trade log also contains 165 positions outside the declared asset universe, while the analysis excludes 217 rows without close dates. Removing the off-universe trades doesn't change the headline results in the robustness check. The opening stop is stored separately and present on all 2,627 rows, so stop coverage isn't a data limitation. The remaining unresolved issue is the 19-position difference between the roster and trade tables, which narrows claims that depend on the exact sample

**6. Limited generalisation and effective sample**

The 2,627 positions don't equal 2,627 independent decisions. Agents held up to five positions concurrently, two parallel runs began in May, and 31% of cross-run positions overlapped in the same symbol and direction. Agents were also inactive during different parts of the window, so some comparisons cover different sub-periods.

The study covers one 6.5-month market sequence, one vendor, and one execution environment. It doesn't test conditions such as front-running, congestion, information leakage, or a participant adapting to the agents' trades. The findings describe how discipline behaved across changing regimes in this deployment snapshot. They don't establish that the same pattern will hold across every market cycle, venue, or live environment.

**Disclaimer: Hygentra warns against treating this research report and its contents as proof that Hygentra's AI agents can consistently outperform markets. See Hygentra.ai or the latest Terms of Service, including but not limited to: "Historical, simulated, recent, or "verified" performance is descriptive only and does not guarantee future results." Nature of the Services as "not personalised investment advice." Agentic Trading and delegated permissions stating setup is non-custodial, and funds stay in the trader's own account, and Risk Disclosure of high risk assets.**

# 8. Related work

**Longitudinal evidence of human trading discipline.** Naja (2025) *27 Years of Blood, Sweat and Lost Billions* prior 2025 longitudinal retail-trading study,, analyses 8 million trader profiles and 295 million trades across 27 years. The study reports that 85% of failed accounts followed the same four-phase pattern: cautious success, overconfidence, catastrophic loss, and terminal decline. It also identifies the behaviors that accompany that breakdown, including increasing position sizes, widening stop-losses, loss chasing, and abandoning established risk rules. This research provides a long-run human baseline for the discipline question. It motivates AI intervention at the execution stage, but it doesn't test autonomous agents or show that AI intervention caused better outcomes. This report examines the agent-level question: whether four common strategy types follow their declared rules, and whether that discipline holds as market regimes change. It tests that question across 2,627 closed positions (2,144 simulated against live prices, 483 real orders on Hyperliquid) from one deployment snapshot.

**Human-defined constraints and accountability.** Jalali and Rafiei (2026) *Process Constitutions and Process Stewards: Towards the Next Generation of BPM for Agentic Organizations* *arXiv:2610.05942v1 [cs.AI],* propose a Process Constitution that defines permitted agent behaviour and a Process Steward that interprets and enforces those boundaries. Their framework reserves constitutional authorship and amendment for human stakeholders, linking agent autonomy to explicit constraints and accountability. Our study provides adjacent empirical evidence from trading agents operating within human-designed strategies and execution controls. We examine whether those controls held as market conditions changed and document exceptions, without implementing or evaluating a Process Steward.

**Execution-policy failures and attribution.** Ardaiz, Budati and Habibnia (2026) *Mixture-of-Experts for Cryptocurrency Order Execution: Training Stability, Tail Risk, and Failure Modes,* *arXiv:2610.03369v1 [q-fin.TR],*evaluate reinforcement-learning policies for cryptocurrency order execution across repeated training runs. Their experiments distinguish mean execution performance, policy collapse and tail risk, showing that apparent benefits from expert partitioning can reflect interactions within the training specification rather than an intrinsic architectural advantage. This motivates careful attribution of observed agent behaviour to the surrounding system. Our study examines a related question at execution time: whether trading agents followed declared risk rules within a human-designed control framework. We assess adherence separately from profitability and document control failures, rather than comparing training architectures.

**Deployment-scale audits of agent performance.** Yu, Zhao and Sui (2026) *Paper Agents, Paper Gains: An Empirical Analysis of DeFi Investment Agents,* *arXiv:2605.29174 [cs.AI]* is the closest prior work and the study this one is designed to complement. Working from on-chain data, they measure eleven agent treasuries and 925,323 token holders, and find a category in which treasury performance does not reliably pass through to the people who funded it, valuations run above 10,000× assets under management where established DeFi protocols sit below 1×, and autonomy can't be verified from public data at all — a wallet reveals what was executed, not what initiated it. Yu et al. assess deployed agents from the outside using daily balance snapshots. Their data show treasury outcomes but provide limited visibility into individual trades, decision-making and rule adherence. Our study provides a robust, reproducible framework for auditing rule adherence across four agents using 2,627 closed-position records, a documented autonomy boundary and a human-intervention log. The dataset is mostly simulated, with 18% of trades executed live on Hyperliquid. Nine audit tests examine whether agents followed their declared rules and whether asset selection explains their results. Our design makes agent behavior directly testable across changing market conditions, while treating profitability as an outcome of the observed period. Our **Sections 4.3–4.5** addresses their second criterion at trade level, and our **Section 3.2** documents the autonomy boundary their first criterion asks for.

**Whether model capability is the constraint.** Chen et al. *(2025) StockBench: Can LLM Agents Trade Stocks Profitably In Real-world Markets?, arXiv:2510.02209 [cs.LG]* found that most LLM agents fail to outperform buy-and-hold on DJIA constituents over a four-month horizon. Fan et al. (*AI-Trader, arXiv:2512.10971, [q-fin.CP]*) extend the evaluation to US stocks, A-shares and cryptocurrencies and conclude that general intelligence doesn't translate into trading capability. Qian et al. (*When Agents Trade, arXiv:2510.11695, [cs.CL]*) report that system-level design choices: multi-agent collaboration, tool use, memory; influence profitability more than the choice of underlying model. *Luo, Y., et al. (2025). LLM-Powered Multi-Agent System for Automated Crypto Portfolio Management, arXiv:2501.00826 [q-fin.TR]* finds a multi-agent crypto portfolio system outperforming single-agent baselines on large-cap tokens. The paper has a substantial 2026 revision; identify the version actually used.

**Live-market evaluation of LLM trading.** Nof1's *Exploring the Limits of Large Language Models as Quant Traders* introduces Alpha Arena, a live benchmark that gave six leading LLMs $10,000 each to trade crypto perpetuals on Hyperliquid with zero human intervention. Each model received the same prompt, harness, and numerical market data, with actions limited to entering long, entering short, holding, or closing. Nof1 reports meaningful differences in risk appetite, position sizing, holding time, trade frequency, exit-plan design, and sensitivity to small prompt changes. It also documents contradictions in self-authored plans and rule-gaming under one harness variant. The live-capital design exposes fees, execution conditions, adverse selection, and counterparties that paper trading can't reproduce. Nof1 also cautions that Season 1 was a short, prompt-sensitive experiment with limited statistical power, so it doesn't establish a permanent ranking of models. Alpha Arena complements this paper by testing zero-shot model behavior under a common live harness. Hygentra's study examines four common strategy classes within a human-designed framework and asks whether declared rules and risk discipline hold as market regimes change. Together, the studies separate model behavior from the discipline built into the wider trading system.

**Industry evidence on regime-aware AI allocation.** Steve Goldstein's MarketWatch article, *"There's good and bad news if you want to copy JPMorgan's new AI agent that outperforms the 60/40 portfolio"* (July 13, 2026), reports that JPMorgan strategists led by Thomas Salopek designed an AI agent to classify market conditions as Goldilocks, reflation, stagflation, or risk-off. The agent then applied a fixed asset-allocation strategy across equities, bonds, credit, commodities, and risk exposures, with the stated aim of outperforming a standard 60/40 portfolio with less risk. MarketWatch recreated the process using lagged economic and market data, along with two Anthropic and two OpenAI models. The reported model portfolios outperformed the simple 60/40 benchmark, although the models produced similar rather than identical results and sometimes disagreed with themselves when prompted repeatedly. The article also raises a data-leakage concern: date-anonymised prompts and lagged inputs don't prevent models from recalling recognisable historical episodes such as 2008 or Covid. For this report, the relevance is methodological. The JPMorgan example shows how an AI system can classify regimes and apply a different allocation rule to each one. Hygentra's study asks a related question at trade level: whether four common trading strategies maintain discipline as market conditions change. The JPMorgan example remains a backtest and model exercise, so it doesn't establish live execution quality or rule adherence. This body of work evaluates agents on controlled tasks and largely agrees that the model isn't the differentiator. It leaves open what is. Our study is the deployment-scale version of that question: we hold the environment, the asset universe and the execution engine fixed across four agents and ask which observable behaviors track the profitability dispersion that results.

**Infrastructure controls and outcome evaluation.** Mudgal, Zhao, Zhang, Olsen, Miller, Chu and Blanken (2026), *MemLeak: Cross-User Semantic Leakage in Multi-Tenant AI Agent Memory, arXiv:2610.04195v1 [cs.AI]*, examine cross-user memory leakage in multi-tenant AI agents. Their controlled experiments show that responses incorporating another user's memories can still receive favorable helpfulness scores. Among three architectural mitigations tested, only hard post-retrieval ownership gating consistently restored the clean baseline across two generation models. This provides an adjacent example of why favorable outcome evaluations do not establish compliance with system boundaries. Our study examines this distinction in trading by assessing adherence to declared risk rules separately from profitability and documenting execution-control failures.

**Structural constraints on agent trading.** Fábrega, A., et al. (2026). *The CoinAlg Bind: Profitability-Fairness Tradeoffs in Collective Investment Algorithms, arXiv:2601.00523 [cs.GT]* prove that collective investment algorithms can't simultaneously guarantee economic fairness and avoid ceding profit to arbitrage, and note that agents with publicly observable wallets are structurally exposed to front-running. Under the paper's formal assumptions; wallet visibility alone is not the complete condition. That result bears directly on how our findings should be read. Of the 2,627 positions analysed, 2,144 were simulated against live venue prices and 483 involved real orders on Hyperliquid **(Section 2.5)**. The simulated positions did not interact with a live order book; the live orders were exposed to actual execution conditions. We did not separately measure front-running, information leakage or adversarial responses to those orders. The combined results therefore document observed trading behavior, but do not establish resistance to exploitation or predict future returns. Trading-system decomposition The expectancy decomposition used in **Section 4.3 Attribution drivers: hit rate vs payoff ratio** — hit rate × average win − loss rate × average loss — is standard in the practitioner literature on system evaluation. Expectancy follows directly from the definitions in **Section 3.2. Per-agent metrics.** The regime labelling in **Section 3.3 Regime labels** follows the general practice of using only information available at the labelling date to avoid look-ahead bias.

# 9. Conclusion

This study examined 2,627 closed paper positions from four AI trading agents over six and a half months, to answer five detailed questions that have been distilled into three questions.

**What can an investor verify from an agent's trade record?**
An investor can check whether the agent followed its stated trading rules and identify where it fell short. Readers can rerun the trade-log tests from the published trade snapshot. Checks across market conditions also need the published Bitcoin price data and market-condition labels. To count stop breaches, readers should apply the allowance for small execution differences stated in **Section 3.5.** The agents held 97.5% of gross loss inside their declared stops, kept declared risk stable as volatility rose and fell, did not widen risk after a loss, did not trade more after a loss, and respected their position ceiling in 23 of 24 runs. We ran these checks using the trade snapshot and the supporting inputs described above. The single breach was found the same way.

**What separated the agents' results?**
All four agents' declared rules held meaning the controls kept losses contained. The model set each stop, target and size within a human-designed framework, and the execution layer enforced them. The ranking came from the margin between how often an agent won and how often its own stop and target required it to win. Payoff ratio, loss size and asset mix didn't explain it after reweighting and statistical controls.

**Did the same agent remain most profitable as market conditions changed?**
No particular agent owned every market regime. Volatility moved results more than directional alignment, and weekly leadership rotated. Behavior was consistent. Leadership was not.

What that adds up to matters most for the investor who has never had access to any of it. The thing separating a professional desk from a person trading at home was never mainly intelligence or information. It's the control layer: stops that hold, risk that doesn't grow after a loss, a ceiling that is enforced rather than intended. That layer used to require an institution to pay for it. In this fleet it ran by default, and it held through a 29% drawdown and the recovery that followed, with two documented exceptions: one run above the position limit for 28 days (Feb–Mar), and 13 days in August when engine stop/target execution was disabled.

We aren't asking anyone to take that on faith, and we aren't asking anyone to build an agent. **Section 6** sets out the checks in the order we would run them: establish which trade instructions come from the model and which controls come from humans, request every closed trade rather than a summary, then test loss containment, post-loss behavior, asset concentration, break-even hit rate after costs, and results by market condition and by week. None of it requires a trading background. These checks use the published trade snapshot, the BTC price and label files, and the stop-breach tolerance in **Section 3.5.** Investors need access to those inputs to verify the results.

The evidence base will improve from here, through longer live records, account-level sizing, realistic costs, more market cycles and more independent vendors. It will improve faster if traders ask for those things, because builders supply what buyers demand, and that part is within reach of everyone. You don't need capital, a trading background, or a view on where the market goes next to ask for a trade record and check it. Rule-based agent systems don't guarantee returns. Agents can give self-directed traders a professional process they can use and inspect, which hasn't previously been available at retail. The only thing it costs is the willingness for traders to try out agents for investing.

# Acknowledgments and Disclosure of Funding

The authors thank colleagues at Hygentra for discussions and feedback on drafts. The research was conducted as part of Hygentra's routine agent performance evaluation of multi-agent architecture, reliability, observability, ongoing training process and was funded internally. Large language models assisted with drafting and editing the text. The authors designed the method, ran the experiments and verified every numeric claim against the underlying run data.